\def\year{2018}\relax
\documentclass[letterpaper]{article} %DO NOT CHANGE THIS
\usepackage{aaai18}  %Required
\usepackage{times}  %Required
\usepackage{helvet}  %Required
\usepackage{courier}  %Required
\usepackage{url}  %Required
\usepackage{graphicx}  %Required
\usepackage{amsmath}
\usepackage{xcolor,etoolbox}

\newtoggle{paperfinal}
\settoggle{paperfinal}{true}

\begin{document}
\title{Approximating meta-heuristics with homotopic recurrent  neural networks}

\iftoggle{paperfinal}{
\author{Alessandro Bay\\
Cortexica Vision Systems Ltd.\\
London, UK
\And
Biswa Sengupta\\
Imperial College London\\
Cortexica Vision Systems Ltd.\\
London, UK
}
%PDF Info Is Required:
  \pdfinfo{
/Title (Approximating meta-heuristics with homotopic deep recurrent  neural networks)
/Author (Alessandro Bay and Biswa Sengupta)}
}{
\author{}
%PDF Info Is Required:
  \pdfinfo{
/Title (LSTM for inference)
/Author (....)}
}

\maketitle

\begin{abstract}
Much combinatorial optimisation problems constitute a non-polynomial (NP)  hard optimisation problem, i.e., they can not be solved in polynomial time. One such problem is finding the shortest route between two nodes on a graph.  Meta-heuristic algorithms such as $A^{*}$ along with mixed-integer programming (MIP) methods are often employed for these problems. Our work demonstrates that it is possible to approximate solutions generated by a meta-heuristic algorithm using a deep recurrent neural network. We compare different methodologies based on reinforcement learning (RL) and recurrent neural networks (RNN) to gauge their respective quality of approximation. We show the viability of recurrent neural network solutions on a graph that has over  300 nodes and argue that a sequence-to-sequence network rather than other recurrent networks has improved approximation quality. Additionally, we argue that homotopy continuation -- that increases chances of hitting an extremum -- further improves the estimate generated by a vanilla RNN.
\end{abstract}

\section{Introduction}
\label{sec:intro}

NP (non-deterministic polynomial time) hard problems are the mainstay of some fields, from problems in graph theory (routing/scheduling, etc.) to mathematical programming (knapsack, 3d-matching, etc.). A widely studied NP-hard problem is the Travelling Salesman Problem (TSP) and its derivatives that include finding the shortest routes between two nodes of a graph. Apart from mixed-integer programming (MIP), solutions of such problems are well approximated by meta-heuristic formulations such as tabu search, simulated annealing, genetic algorithms and evolutionary optimisations. Of notable mention are the Dantzig-Fulkerson-Johnson algorithm \cite{Dantzig1954}, branch-and-cut algorithms \cite{Naddef2001}, neural networks \cite{Ali1993}, etc.

The shortest-path problem is central to many real-life scenarios, from inventory delivery (courier, food, vehicles, etc.) to laying out circuitry on a printed circuit board. This problem has evolved from a discrete integer programming (bottom-up) problem to a data-based continuous optimisation problem (top-down). Some studies have used specialised reinforcement learning (RL) algorithms -- such as q-learning -- to approximate the optimal solutions for the shortest path problem \cite{Boyan1994}. Along with the resurgence of deep neural networks (DNN), techniques that merge the scalability of deep neural networks and the theoretical framework of Markov Decision Processes (MDP) have emerged (\textit{aka. }deep reinforcement learning (DRL)) \cite{Bello2016,Dai2017}. The intrinsic non-convexity of the loss function means both DRL and DNN struggle to find the global optimisers of the loss function. This becomes more of a problem as the graph size increases. 

For a deep neural network, finding the shortest routes between two points can be framed as a sequence learning problem. Indeed, in this paper, we show how synthetic routes generated by an $A^{*}$ algorithm can be approximated using recurrent neural networks. With the rise of the Seq2Seq (sequence to sequence) architecture,  recurrent neural networks based on Long Short Term Memory (LSTMs), Gated Recurrent Units (GRUs) and others can be readily used as a sub-module. In this paper, we concentrate on three shortest path finding algorithms on a reasonably sized graph of more than  300 nodes. We use -- (a) meta-heuristics based $A^{*}$ algorithm \cite{Hart1968}, (b) q-learning based Q-routing algorithm \cite{Boyan1994} and (c) a Seq2Seq recurrent neural network \cite{Sutskever2014}, amongst other vanilla recurrent architectures. Moving on, we argue that the value of extremums generated by the RNN could be further improved via the homotopic continuation of the neural network's loss function. 

\section{Methods}
\label{sec:methods}

\subsection{Datasets}

The graph is based on the road network of Minnesota\footnote{\url{https://www.cs.purdue.edu/homes/dgleich/packages/matlab_bgl}}. Each node represents the intersections of roads while the edges represent the road that connects the two points of intersection. Specifically, the graph we considered has 376 nodes and  455 edges, as we constrained the coordinates of the nodes to be in the range $[-97,-94]$ for the longitude and $[46,49]$ for the latitude, instead of the full extent of the graph, i.e., a longitude of $[-97,-89]$ and  a latitude of $[43,49]$, with a total number of 2,642 nodes.

\subsection{Algorithms}

\subsubsection{The $A^{*}$ meta-heuristics}
\label{sec:astar}
\begin{figure}[!h]
\centering{
\includegraphics[scale=0.6]{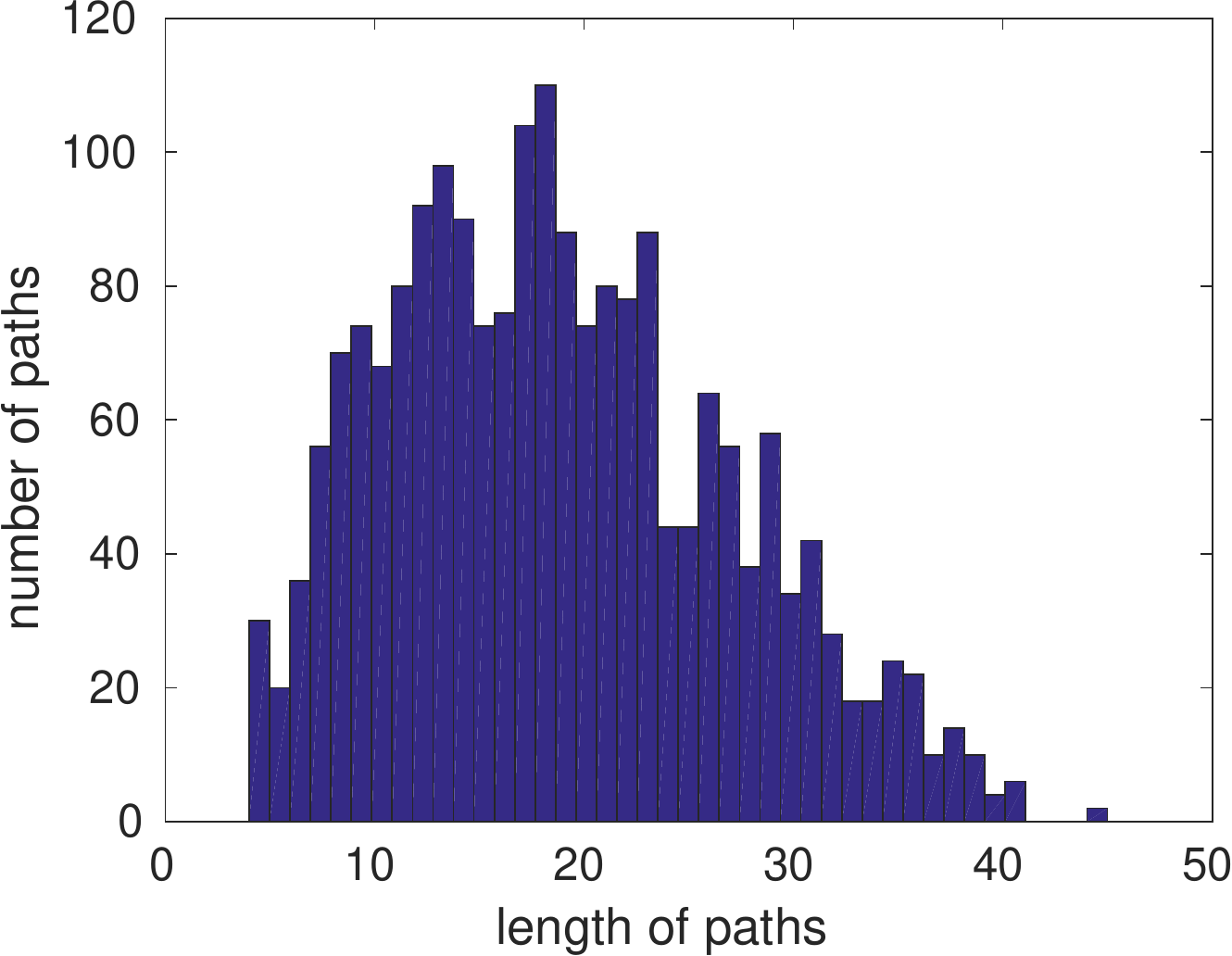}
\caption{\textbf{Distribution of paths lengths.} After selecting two nodes uniformly at random, we compute the shortest paths using the $A^*$ algorithm. The average path length is 19 hops.}
\label{fig:histLength}}
\end{figure}

The $A^{*}$ algorithm is a best-first search algorithm wherein it searches amongst all of the possible paths that yield the smallest cost. This cost function is made up of two parts -- particularly, each iteration of the algorithm consists of first evaluating the distance travelled or time expended from the start node to the current node. The second part of the cost function is a heuristic that estimates the cost of the cheapest path from the current node to the goal. Without the heuristic part, this algorithm operationalises the Dijkstra's algorithm \cite{Dijkstra1959}. There are many variants of $A^{*}$; in our experiments, we use the vanilla $A^{*}$ with a heuristic based on the Euclidean distance. Other variants such as Anytime Repairing $A^{*}$ has been shown to give superior performance \cite{Likhachev2004}. $A^{*}$ generated paths between two randomly selected nodes are calculated. On an average, the paths are {19} hops long and follow the distribution represented by histograms in Figure \ref{fig:histLength}.

\subsubsection{De-centralised Q-routing}
\label{sec:qrouting}
Q-Learning is an Off-Policy algorithm for temporal difference learning. \citeauthor{Boyan1994} (\citeyear{Boyan1994}) have formulated the Q-routing algorithm by building a routing table based on node distance/time (q-values). This algorithm utilises q-learning wherein the nodes that are in the neighbourhood of the current node communicate the expected future waiting time. In Q-routing, every source node $x$ has to choose the interim node that leads it to the destination node $d$. Q-learning enables us to learn the expected travel time to $d$ for each of the possible interim node $y$. A q-table $Q^{x}$ is created for each node $x$ that is updated at discrete intervals $t$ as,

\begin{eqnarray}
Q_t^x\left( {d,y} \right) &=& \left( {1 - \alpha } \right)Q_{t - 1}^x\left( {d,y} \right) \nonumber \\
&+& \alpha \left( {b_t^x + \mathop {\min }\limits_z Q_{t - 1}^y\left( {d,z} \right)} \right).
\end{eqnarray}

Here, $0 < \alpha  < 1$ is the learning rate and $b_t$ is the time spent at node $x$ before being sent off at time $t$. Q-learning can thus estimate the value/cost (for Q-routing it is the estimated transit time from the current node $x$ to destination $d$ via node $z$) function $V$ for being in state $d$ as $V = \min {Q^x}\left( {d,z} \right)$. A greedy policy is then executed to transact optimally.

\subsubsection{Recurrent deep networks}

\begin{figure}[h]
\centering
\includegraphics[scale=0.3]{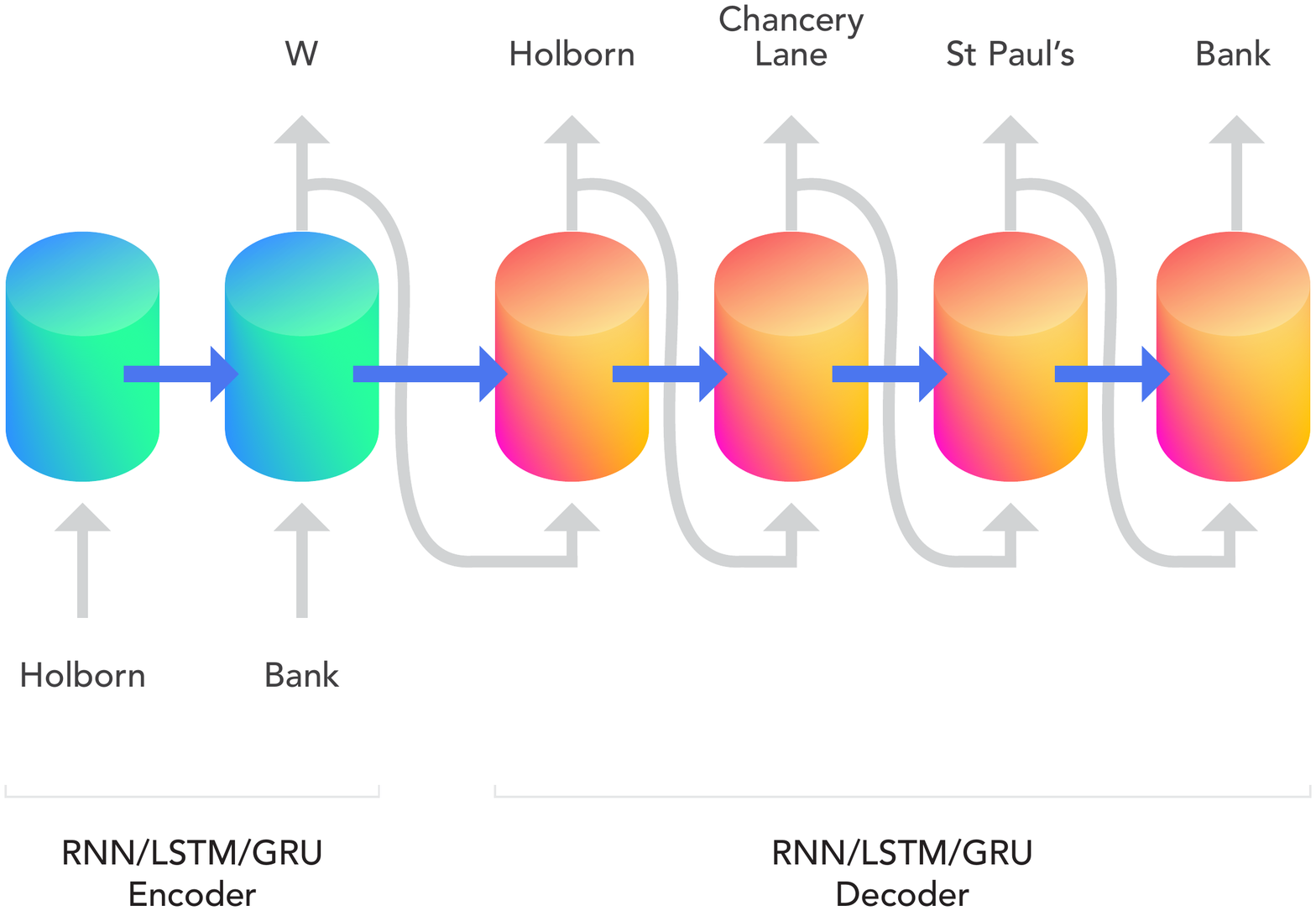}
\caption{\textbf{Sequence-to-Sequence architecture for approximating the $A^{*}$ meta-heuristics.} Here, the first two modules on the left are the encoder while the last four represent the decoded output, representing the shortest route between Holborn and Bank. The network is trained using shortest route snippets that have been generated using an $A^{*}$ algorithm. $W$ represents the context vector.}
\label{fig:seq2seqGraph}
\end{figure}

We utilised a variety of recurrent neural networks for shortest route path predictions:

\begin{itemize}
\item A vanilla RNN \cite{Goodfellow2016}: 
\begin{equation}\label{eq:RNN}
\begin{cases}
h(t) = \tanh(A x(t) + B h(t-1) + b) \\
y(t) = f(C h(t) + c)
\end{cases},
\end{equation}
with weights $\{A,B,b,C,c,h_0=h(0)\}$, which takes as input $x(t)$ the current node, described by a \emph{one-hot} representation, and estimates the following one. The function $f$ can be a $\textrm{softmax}$, since we expect a probability distribution on the following node, or a $\textrm{log softmax}$, that gives more numerical stability during training. During the test phase, we use the predicted node as input for the next time step and compute two paths, one starting from the source and one from the destination, that form an intersection to obtain the shortest path.

\item A \textit{Seq2Seq}-based model \cite{Sutskever2014}: 
We start with the tuple [source, destination] as an input sequence, which is encoded by a vanilla RNN (Figure \ref{fig:seq2seqGraph}). The context vector, i.e., the encoding, is then decoded by another RNN, LSTM or a GRU \cite{Cho2014} to obtain the shortest path connecting the source to the destination.

\item A \textit{Seq2Seq}-based model with attention \cite{Bahdanau2014} enables us to focus on the encoded states with varying extent.

\item A vanilla RNN, trained with homotopy continuation \cite{Vese1999} -- convolution of the loss with a Gaussian kernel, as explained in the following section.

\end{itemize}

\subsection{Homotopy continuation}\label{sec:homotopy}
In algebraic topology, two continuous functions in topological space that can be transformed (deformed) from one to another are known as homotopies. In the context of neural networks, such homotopies enable us to pose different instantiations of the loss function with the aim of obtaining a global minimum. Deep neural networks always have a non-convex loss with no convergence guarantees; this makes learning optimal parameters a numerical exercise. A homotopy continuation allows us to gradually deform the loss function to a non-convex one from another that has a well-defined minimiser. The minimisers from the $i$-th sub-problem are the starting points for the subsequent $(i+1)$-th subproblem.     

Assume that $g(x)=0$ represents a problem with global optimizers whilst $f(x)$  is another function where we have no \textit{a priori} knowledge of the optimizers. Then the homotopy function becomes $h\left( {x,t} \right) \equiv \left( {1 - t} \right)g\left( x \right) + tf\left( x \right):t \in \left[ {0,1} \right]$. The path traced out by the states are governed by $\dot x =  - {\left[ {{J_x}\left( {x,t} \right)} \right]^{ - 1}}\frac{{\partial h}}{{\partial t}}$. If the Jacobian $J$ is of full-rank, one can guarantee that the path between the two functions are continuously differentiable. In our treatise, we use a continuation path defined by the heat equation. In a similar vein to  \citeauthor{Mobahi2016} (\citeyear{Mobahi2016}), we convolve the loss function with an isotropic Gaussian (the Weierstrass transform) of standard deviation $\sigma$; this enables us to obtain objective functions with varying smoothness. $\sigma$ then becomes the continuation parameter for the diffused cost function. We illustrate the basic methodology using a single dimensional function, followed by the diffusion of the RNN's loss function.

\begin{equation}\label{eq:K}
K(x,\sigma) = \frac{1}{\sqrt{2\pi} \sigma} \exp\bigg(\frac{-x^2}{2 \sigma^2}\bigg)
\end{equation}

\begin{figure}[h]
\centering
\includegraphics[scale=0.5]{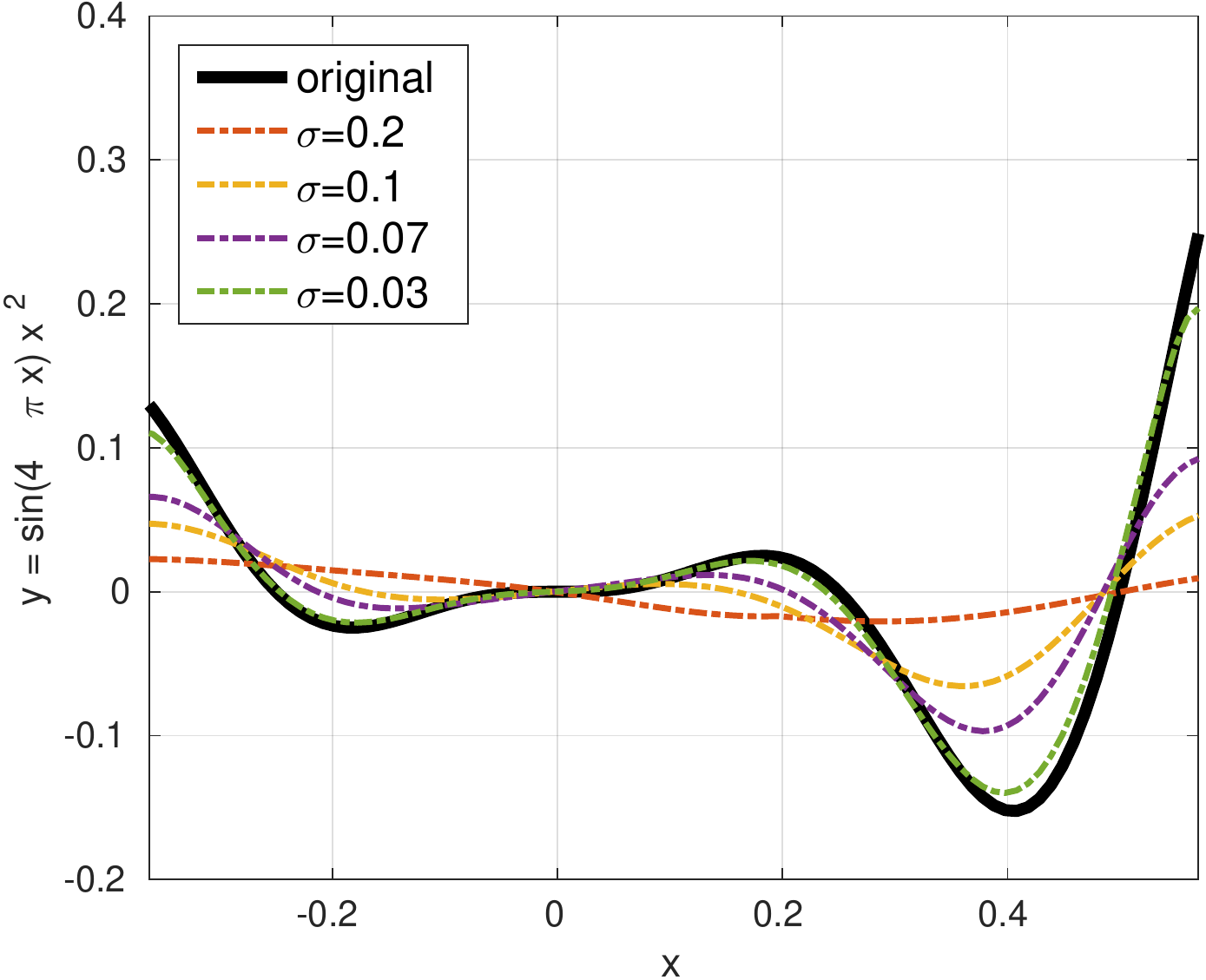}
\caption{\textbf{Example of a diffused cost function.} We choose a cost function equal to $\sin(4\pi x)x^2$, which has both local and global minima for $x \in [-0.4,0.6]$. Applying the Weierstrass transform we obtain smoothed forms for the cost function, which converges to the original cost as $\sigma$ decreases.}
\label{fig:exampleDiffusion}
\end{figure}

As an illustration, we construct a  cost function that has both a local and global minima as well as a saddle point, i.e., $y=\sin(4\pi x)x^2$, constrained to the interval $[-0.4,0.6]$. This function is shown in Figure \ref{fig:exampleDiffusion} as a solid black line, along with its diffused forms, obtained by the convolution with the Gaussian kernel (Eqn. \ref{eq:K}) with standard deviation $\sigma = \{0.2,0.1,0.07,0.03\}$. As one can see, for higher values of $\sigma$ (dashed red line), the smoothed function has only one minimum, which can be easily attained via an arbitrary optimisation method. Then, starting from this solution, we can avoid falling into the local minimum and eventually reach the global extremiser for smaller values of $\sigma$ (other dashed lines). Repeating this until $\sigma$ attains low values allows the diffused function to match the original, thereby obtaining the global minimum.

\subsubsection{Diffused cost function}

In this section, we illustrate the equations leading to the diffused loss function for the RNN. Notably, given a set of $S$ shortest sequences, each of length $T_s$, and a loss function $d$, we can state the minimisation problem, of the cost function \textit{w.r.t.} the weights in a vanilla RNN (Eqn. \ref{eq:RNN}) as,

\begin{align}\label{eq:costConstr}
&\min_{A,B,b,C,c,h_0} \frac{1}{S} \sum_{s=1}^S \sum_{t=1}^{T_s} d(f(z_y(t)),y(t)), \\
\textrm{s.t. } & z_y(t) = C \tanh(z_h(t)) + c \nonumber\\
               & z_h(t) = A x(t) + B h(t-1) + b, \nonumber
\end{align}
where $h(0)=h_0$ is the initial hidden state.

Moreover, if we consider $z_y$ and $z_h$ as independent variables, we can write the following unconstrained Lagrangian,

\begin{align}\label{eq:costUnconstr}
&\min_{A,B,b,C,c,h_0,Z_h,Z_y} \frac{1}{S} \sum_{s=1}^S \sum_{t=1}^{T_s} d(f(z_y(t)),y(t)) \nonumber\\
&+ \lambda \bigg(p(C \tanh(z_h(t)) + c - z_y(t)) \\
&+ p(A x(t) + B h(t-1) + b - z_h(t))\bigg), \nonumber
\end{align}
where all the vectors $z_h(t)$ and $z_y(t)$ are collected in the columns of the matrices $Z_h$ and $Z_y$, respectively, $\lambda$ is a regularization parameter (we will use $\lambda = 0.01$), and $p(\cdot)$ is a penalty function.

Since we consider $f(\cdot) = \textrm{log softmax}(\cdot)$, then the natural choice for the loss $d$ is the negative log-likelihood, while for the penality function $p$, we choose the squared Euclidean norm. Therefore, Eqn. \ref{eq:costUnconstr} now becomes

\begin{align}\label{eq:costUnconstr2}
&\min_{A,B,b,C,c,h_0,Z_h,Z_y} \frac{1}{S} \sum_{s=1}^S \sum_{t=1}^{T_s} -f(z_y(t))^\top y(t) \nonumber\\
&+ \lambda \bigg(\|C \tanh(z_h(t)) + c - z_y(t)\|_2^2 \\
&+ \|A x(t) + B h(t-1) + b - z_h(t)\|_2^2\bigg). \nonumber
\end{align}

Finally, defining the convolution of the non-linear functions $\tanh$ and $f$ with the Gaussian kernel (Eqn. \ref{eq:K}) as
$$\tanh_\sigma(\cdot) := \tanh(\cdot) \ast K(\cdot,\sigma) \qquad f_\sigma(\cdot) := f(\cdot) \ast K(\cdot,\sigma),$$
respectively, we can derive the constrained diffused cost \textit{w.r.t.} the weights of the RNN, as

\begin{align}\label{eq:diffCost}
&\min_{A,B,b,C,c,h_0} \frac{1}{S} \sum_{s=1}^S \sum_{t=1}^{T_s} -f_\sigma(z_y(t))^\top y(t) \nonumber\\
&+ \lambda \bigg(\|C \textrm{diag}(\sqrt{(\tanh^2)_\sigma}(z_h(t)))\|_F^2 \nonumber\\
&- \|C \textrm{diag}(\tanh_\sigma(z_h(t)))\|_F^2 \nonumber\\
&+ \sigma^2 \texttt{outDim}\|\tanh_\sigma(z_h(t))\|_2^2\\
&+ \|B \textrm{diag}(\sqrt{(\tanh^2)_\sigma}(z_h(t-1))) \|_F^2 \nonumber\\
&- \|B \textrm{diag}(\tanh_\sigma(z_h(t-1))) \|_F^2 \nonumber\\
&+ \sigma^2 \texttt{stateDim}\|\tanh_\sigma(z_h(t-1))\|_2^2\bigg), \nonumber\\
\textit{s.t. } & z_y(t) = C \tanh_\sigma(z_h(t)) + c \nonumber\\
               & z_h(t) = A x(t) + B \tanh_\sigma(z_h(t-1)) + b, \nonumber
\end{align}

where $\textrm{diag}(v)$ denotes a diagonal matrix with the vector $v$ on the diagonal, $\|\cdot\|_F$ is the Frobenius norm of a matrix, and $\texttt{stateDim}$ and $\texttt{outDim}$ represent the number of neurons in the hidden state and of the output, respectively. Notice that we have also used the identities that convolution of $\|Af(x)+b\|_2^2$ with $K(x,s)$ is equal to $\|Af_\sigma(x)+b\|_2^2 + \|A \textrm{ diag}(\sqrt{(f^2)_\sigma}(x))\|_F^2 - \|A \textrm{ diag}(f_\sigma(x))\|_F^2$ and $(x^\top y)^2 \ast K(x,\sigma)=(x^\top y)^2 + \sigma^2 \|y\|_2^2$ \cite{Mobahi2016}.
%---------------------------------------------------------------

\subsubsection{Approximation of activation functions}

Another crucial component of this method is the computation of the diffused non-linearities for the considered RNN. Following \cite{Mobahi2016} convolving $\tanh(x) \approx \textrm{erf}(\frac{\sqrt{\pi}}{2} x)$ with $K(x,\sigma)$, we obtain

\begin{align}\label{eq:tanh_s}
\tanh_\sigma(x) &= \tanh(x) \ast K(x,\sigma) \nonumber\\
&= \textrm{erf}\left(\frac{\sqrt{\pi}}{2} x\right) \ast K(x,\sigma) \nonumber\\
&= \textrm{erf}\left(\frac{\sqrt{\pi}}{2} \frac{x}{\sqrt{1+\frac{\pi}{2}\sigma^2}}\right) \nonumber\\
&= \tanh\left(\frac{x}{\sqrt{1+\frac{\pi}{2}\sigma^2}}\right).
\end{align}

However, the output non-linearity that we use is the log softmax, which cannot be analytically convolved. To address this problem, our first attempt was the numerical computation of the convolved function, but in this case, we obtained numerical errors, especially on the boundaries. This makes the approximation less accurate. To sidestep this phenomenon, we consider a linear interpolation $y=mx+q$ of the central interval, obtaining the following slope, which depends on the standard deviation $\sigma$:

\begin{equation}\label{eq:1minus1overPi}
m = \left(1-\frac{1}{\pi}\right)\exp(-\pi\sigma^2)+\frac{1}{\pi},
\end{equation}
and the shift is simply the constant $q=-\log(\sum \exp(x))$.

\begin{figure}[!h]
\centering
\includegraphics[scale=0.5]{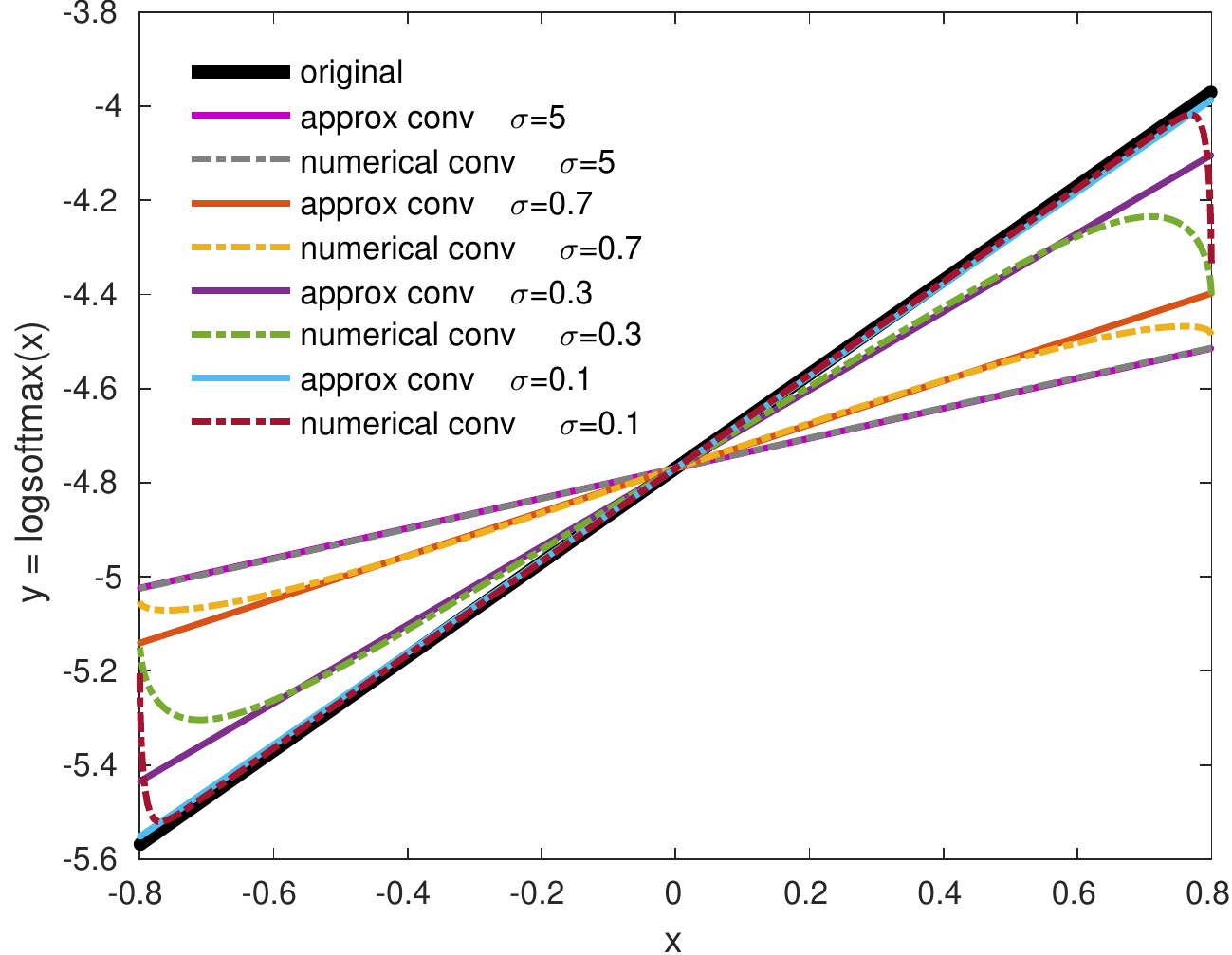}
\caption{\textbf{Approximation for the convolved log softmax.} Since $\textrm{log softmax}$ function cannot be analytically convolved with the Gaussian kernel, we approximate this operation numerically. It is apparent that the quality of the approximation is affected by the numerical errors, especially on the boundaries (dash-dotted lines). Therefore, we constrain the approximation via linear interpolation of the central interval (solid lines) at the edges.}
\label{fig:approxLogsoftmax}
\end{figure}

As we can see in Figure \ref{fig:approxLogsoftmax}, some errors affect the numerical convolution, especially on the boundaries (dash-dotted lines). Thus, a linear equation with slope as in Eqn. \ref{eq:1minus1overPi} is more accurate and converges to the original logsoftmax function (solid black line) as $\sigma$ becomes smaller (in Figure \ref{fig:approxLogsoftmax} we illustrate the results for $\sigma=\{5,0.7,0.3,0.1\}$).

Table \ref{table:diffusedActivations} illustrates the diffused forms of the most popular activation functions.

\begin{table}[!h]
\centering{
\resizebox{1\columnwidth}{!}{
\begin{tabular}{l c c c}
\hline
function & original & diffused \\
\hline
error  & $\textrm{erf}(\alpha x)$ & $\textrm{erf}\left(\frac{\alpha x}{\sqrt{1+2(\alpha\sigma)^2}}\right)$\\
tanh & $\tanh(x)$ & $\tanh\left(\frac{x}{\sqrt{1+\frac{\pi}{2}\sigma^2}}\right)$ \\
sign & $ \begin{cases} +1 \quad \textrm{if } x>0\\ \;\,\,\,0 \quad \textrm{if } x=0 \\ -1 \quad \textrm{if } x<0\end{cases} $ & $\textrm{erf}\left(\frac{x}{\sqrt{2}\sigma}\right)$ \\
relu & $\max(x,0)$ & $\frac{\sigma}{\sqrt{2\pi}}\exp\left(\frac{-x^2}{2\sigma^2}\right)+\frac{1}{2}x\left(1+\textrm{erf}\left(\frac{x}{\sqrt{2}\sigma}\right)\right)$ \\
logsoftmax & $x-\log\bigg(\sum \exp(x)\bigg)$ &  $\bigg(\left(1-\frac{1}{\pi}\right)\exp\left(-\pi\sigma^2\right)+\frac{1}{\pi}\bigg) x - \log(\sum(\exp(x)))$ \\ 
\hline
\end{tabular}}
\caption{\textbf{List of diffused forms (Weierstrass transform).} We report the most popular non-linear activation functions along with their diffused form. This is obtained by convolving the function with the heat kernel $K(x,\sigma)$. This table extends the work in \cite{Mobahi2016} by an analytic approximation of the  log softmax function.  }
\label{table:diffusedActivations}}
\end{table}

\section{Results}

For the graph of Minnesota with 376 nodes and  455 edges, we generated 3,000 shortest routes between two randomly picked nodes using the $A^*$ algorithm. We used these routes as the training set for the Q-routing and the RNN algorithms using a 67-33\% training-test splits. 

For the Q-routing algorithm, we set the learning coefficient $\alpha=0.5$ and use 1,000 iterations for each path to update the q-table and enable the algorithm to converge. We obtain an accuracy on the shortest paths of 70\% for the test data-set, and 97\% of the test paths reach the destination, albeit they are not necessarily the shortest ones. 

On the other hand, we choose a hidden state with 256 units for the RNN-based methods and run the training for 200 epochs updating the parameters with an Adam optimisation scheme \cite{Kingma2014} with parameters $\beta_1=0.9$ and $\beta_2=0.999$, starting from a learning rate equal to 1e-3.

\begin{table}[!h]
\centering{
\begin{tabular}{l c c c}
\hline
method & shortest & successful \\
\hline 
Q-routing & $70\%$ & $97\%$ \\
RNN$\_$vanilla & $33\%$ & $33\%$ \\
RNN$\_$intersect & $60\%$ & $98\%$ \\
RNN$\_$withDiff & $63\%$ & $\textbf{99\%}$ \\
Seq2Seq$\_$noAttn & $40\%$ & $40\%$ \\
Seq2Seq$\_$withAttn & $73\%$ & $73\%$ \\
%seq2seqGRU$\_$withAttn & $73\%$ & $73\%$ \\
Seq2Seq$\_$intersect$\_$noAttn & $59\%$ & $83\%$ \\
Seq2Seq$\_$intersect$\_$withAttn & $\textbf{78\%}$ & $88\%$ \\
\hline
\end{tabular}
\caption{\textbf{Results on the Minnesota graph.} Percentage of shortest path and successful paths (that are not necessarily shortest) are shown for a RL algorithm (i.e., Q-routing) and a wide-variety of RNNs, where  \emph{intersect}  means the computation of the shortest path as intersection of the predicted path and the one obtained from the destination to the source node. All scores are relative to an $A^{*}$ algorithm, that achieves a shortest path score of 100\%.}
\label{table:results}}
\end{table}

The prediction accuracy on the test data-set is shown in Table \ref{table:results}. A vanilla RNN predicting the next node gives poor results; especially we notice that most of the time it cannot even predict a path from the source to the destination node. Nevertheless, on occasions when it reaches the destination, the path is always the shortest. A first attempt to improve its performance was simply to compute two paths -- one from the source to the destination node and another one from the destination to the source, and intersecting the two paths to obtain a route between the source and the destination. Such a scheme improved the performance to an accuracy equal to 98\% for paths linking source and destination and to 60\% for shortest ones. Then, we train a vanilla RNN through an homotopy continuation method (see Eqn.\ \ref{eq:diffCost}) varying $\sigma \in \{30,5,1,0.0001\}$, which gives a further improvement (i.e., 63\% and 99\%) and converges after only 80 epochs (instead of 200 epochs). 

Alternatively, we can also alter the architecture of the RNN, getting inspiration from the recent success of the sequence-to-sequence model \cite{Sutskever2014}. We involve it, encoding the [source, destination] tuple via a vanilla RNN and then decoding the context vector into the shortest path sequence. We implemented the decoder in two different ways: it can be either a vanilla RNN or a GRU, or a vanilla RNN or a GRU with attention \cite{Bahdanau2014}. In particular, by embedding attention, we can outperform the accuracy of the Q-routing algorithm on the shortest paths (73\%), especially if we compute two paths and intersect them (78\%).

Therefore, the most accurate algorithm for finding the shortest path is the Seq2Seq model with attention, where during the test phase we evaluate the intersection between two paths: one from the source to the destination node and another from the destination to the source node. On the other hand, a vanilla RNN, when trained with diffusion, computes successful paths from source to destination, albeit they are not necessarily the shortest ones. This suggests that further improvements can be obtained by training a Seq2Seq model with diffusion.

\section{Discussion}

In this paper, we illustrate that recurrent neural networks have the fidelity in approximating routes generated from an $A^*$ algorithm. As the node size increases, Seq2Seq models have increased fidelity compared to vanilla recurrent neural networks or for that matter a vanilla q-learning based routing algorithm. Our work is yet another testament to the utility that deep recurrent networks have in approximating solutions of NP hard problems.  

When parameter evolution is constrained to evolve according to a heat equation, it has resulted in superior posterior estimates when utilised in a Bayesian belief updating scheme \cite{Cooray2017}. Similarly, combining annealed importance sampling with Riemannian samplers have also shown promise in producing better posterior distributions \cite{Penny2016}. Homotopy continuation provides a rigorous theoretical foundation for the afore mentioned results. It is well-known that the convex envelope of a non-convex loss function can be obtained using an evolution equation, i.e., a (non-linear) partial differential equation (PDE) \cite{Vese1999}. The heat equation simply provides an affine solution to this non-linear PDE. One important direction for future work is, therefore, tackling the original nonlinear PDE with computationally efficient algebraic or geometric multi-grid methods \cite{Heroux2012,Sundar2012}. 

We have used Q-routing as a benchmark to learn the action-value pair that gives us the expected utility of taking a prescribed action; in the last few years, many other architectures have evolved such as the deep-Q-network  \cite{Mnih2015}, duelling architecture \cite{Wang2015}, etc. -- it remains to be seen how such deep reinforcement learning architectures cope with the problem at hand. Another vital direction to pursue is the scalability of inverse reinforcement learning algorithms \cite{Ziebart2008} wherein given the policy and the system dynamics the aim is to recover the reward function.  

An important issue arising from using recurrent neural networks for computing shortest path is to memorise long sequences, such as routes that range hundreds of nodes. LSTM \cite{Hochreiter1997} alleviated some of the problems; the other proposition has been to use the second-order geometry as has been long proposed \cite{LeCun2012}. Other efforts have gone towards using evolutionary optimisation algorithms such as CoDeepNEAT for finessing the neural network architecture \cite{Miikkulainen2017}. Similarly, Neural Turing Machines \cite{Graves2014} are augmented RNNs that have a (differentiable) external memory that they can selectively read/write, enable the network to store the latent structure of the sequence. Attention networks, as has been used here, enable the RNNs to attend to snippets of their inputs (\textit{cf.} in conversational modelling \cite{Vinyals2015}). Nevertheless, for the long-term dependencies that shortest paths in large graphs have, these methods are steps towards alleviating the central problem of controlling spectral radius in recurrent neural networks.

The Q-routing algorithm reduces the search-space by constraining the search to the connected neighbours of the current node. Whereas, the RNN variants have a large state-space to explore. The Seq2Seq architecture, therefore, has at least two orders of magnitude increase in its exploration space. In future, the accuracy of the RNN variants can be increased by constraining the search space to the neighbourhood of the current node. 

In a real-world scenario, computing the shortest path between two nodes of a graph is also constrained by the computational time of the algorithm. Thus, the deep recurrent networks, as utilised here, are faced with two objectives -- first, to reduce the prediction error and second to reduce the computational effort. In general, it is quite uncommon that an obtained solution can be obtained that minimises both objectives. In lieu, one can concentrate effort on achieving solutions on the Pareto front. Utilising, Bayesian optimisation \cite{Brochu2010} for such multi-objective optimisation problem is a path forward for research that is not only theoretically illuminating but also commercially useful.   

\iftoggle{paperfinal}{
\section{ Acknowledgments}
BS is thankful to the Issac Newton Institute for Mathematical Sciences for hosting him during the ``Periodic, Almost-periodic, and Random Operators" workshop. 
}

\bibliography{LSTM_inference_NPhard}

\begin{thebibliography}{}

\bibitem[\protect\citeauthoryear{Ali and Kamoun}{1993}]{Ali1993}
Ali, M. K.~M., and Kamoun, F.
\newblock 1993.
\newblock Neural networks for shortest path computation and routing in computer
  networks.
\newblock {\em IEEE Transactions on Neural Networks} 4(6):941--954.

\bibitem[\protect\citeauthoryear{Bahdanau, Cho, and
  Bengio}{2014}]{Bahdanau2014}
Bahdanau, D.; Cho, K.; and Bengio, Y.
\newblock 2014.
\newblock Neural machine translation by jointly learning to align and
  translate.
\newblock {\em arXiv preprint arXiv:1409.0473}.

\bibitem[\protect\citeauthoryear{Bello \bgroup et al\mbox.\egroup
  }{2016}]{Bello2016}
Bello, I.; Pham, H.; Le, Q.~V.; Norouzi, M.; and Bengio, S.
\newblock 2016.
\newblock Neural combinatorial optimization with reinforcement learning.
\newblock {\em arXiv preprint arXiv:1611.09940}.

\bibitem[\protect\citeauthoryear{Boyan and Littman}{1994}]{Boyan1994}
Boyan, J.~A., and Littman, M.~L.
\newblock 1994.
\newblock Packet routing in dynamically changing networks: A reinforcement
  learning approach.
\newblock In {\em Advances in neural information processing systems},
  671--678.

\bibitem[\protect\citeauthoryear{Brochu, Cora, and
  De~Freitas}{2010}]{Brochu2010}
Brochu, E.; Cora, V.~M.; and De~Freitas, N.
\newblock 2010.
\newblock A tutorial on {Bayesian optimization} of expensive cost functions,
  with application to active user modeling and hierarchical reinforcement
  learning.
\newblock {\em arXiv preprint arXiv:1012.2599}.

\bibitem[\protect\citeauthoryear{Cho \bgroup et al\mbox.\egroup
  }{2014}]{Cho2014}
Cho, K.; Van~Merri{\"e}nboer, B.; Bahdanau, D.; and Bengio, Y.
\newblock 2014.
\newblock On the properties of neural machine translation: Encoder-decoder
  approaches.
\newblock {\em arXiv preprint arXiv:1409.1259}.

\bibitem[\protect\citeauthoryear{Cooray \bgroup et al\mbox.\egroup
  }{2017}]{Cooray2017}
Cooray, G.~K.; Rosch, R.; Baldeweg, T.; Lemieux, L.; Friston, K.; and Sengupta,
  B.
\newblock 2017.
\newblock Bayesian belief updating of spatiotemporal seizure dynamics.
\newblock In {\em ICML Time Series Workshop}.

\bibitem[\protect\citeauthoryear{Dai \bgroup et al\mbox.\egroup
  }{2017}]{Dai2017}
Dai, H.; Khalil, E.~B.; Zhang, Y.; Dilkina, B.; and Song, L.
\newblock 2017.
\newblock Learning combinatorial optimization algorithms over graphs.
\newblock {\em arXiv preprint arXiv:1704.01665}.

\bibitem[\protect\citeauthoryear{Dantzig, Fulkerson, and
  Johnson}{1954}]{Dantzig1954}
Dantzig, G.; Fulkerson, R.; and Johnson, S.
\newblock 1954.
\newblock Solution of a large-scale traveling-salesman problem.
\newblock {\em Operations Research} 2:393--410.

\bibitem[\protect\citeauthoryear{Dijkstra}{1959}]{Dijkstra1959}
Dijkstra, E.~W.
\newblock 1959.
\newblock A note on two problems in connexion with graphs.
\newblock {\em Numerische mathematik} 1(1):269--271.

\bibitem[\protect\citeauthoryear{Goodfellow, Bengio, and
  Courville}{2016}]{Goodfellow2016}
Goodfellow, I.; Bengio, Y.; and Courville, A.
\newblock 2016.
\newblock {\em Deep Learning}.
\newblock MIT Press.

\bibitem[\protect\citeauthoryear{Graves, Wayne, and
  Danihelka}{2014}]{Graves2014}
Graves, A.; Wayne, G.; and Danihelka, I.
\newblock 2014.
\newblock Neural turing machines.
\newblock {\em arXiv preprint arXiv:1410.5401}.

\bibitem[\protect\citeauthoryear{Hart, Nilsson, and Raphael}{1968}]{Hart1968}
Hart, P.~E.; Nilsson, N.~J.; and Raphael, B.
\newblock 1968.
\newblock A formal basis for the heuristic determination of minimum cost paths.
\newblock {\em IEEE transactions on Systems Science and Cybernetics}
  4(2):100--107.

\bibitem[\protect\citeauthoryear{Heroux and Willenbring}{2012}]{Heroux2012}
Heroux, M.~A., and Willenbring, J.~M.
\newblock 2012.
\newblock A new overview of the trilinos project.
\newblock {\em Scientific Programming} 20(2):83--88.

\bibitem[\protect\citeauthoryear{Hochreiter and
  Schmidhuber}{1997}]{Hochreiter1997}
Hochreiter, S., and Schmidhuber, J.
\newblock 1997.
\newblock Long short-term memory.
\newblock {\em Neural Comput.} 9(8).

\bibitem[\protect\citeauthoryear{Kingma and Ba}{2014}]{Kingma2014}
Kingma, D., and Ba, J.
\newblock 2014.
\newblock Adam: A method for stochastic optimization.
\newblock {\em arXiv preprint arXiv:1412.6980}.

\bibitem[\protect\citeauthoryear{LeCun \bgroup et al\mbox.\egroup
  }{2012}]{LeCun2012}
LeCun, Y.~A.; Bottou, L.; Orr, G.~B.; and M{\"u}ller, K.-R.
\newblock 2012.
\newblock Efficient backprop.
\newblock In {\em Neural networks: Tricks of the trade}. Springer.
\newblock  9--48.

\bibitem[\protect\citeauthoryear{Likhachev, Gordon, and
  Thrun}{2004}]{Likhachev2004}
Likhachev, M.; Gordon, G.~J.; and Thrun, S.
\newblock 2004.
\newblock {ARA*}: Anytime {A*} with provable bounds on sub-optimality.
\newblock In {\em Advances in Neural Information Processing Systems},
  767--774.

\bibitem[\protect\citeauthoryear{Miikkulainen \bgroup et al\mbox.\egroup
  }{2017}]{Miikkulainen2017}
Miikkulainen, R.; Liang, J.; Meyerson, E.; Rawal, A.; Fink, D.; Francon, O.;
  Raju, B.; Navruzyan, A.; Duffy, N.; and Hodjat, B.
\newblock 2017.
\newblock Evolving deep neural networks.
\newblock {\em arXiv preprint arXiv:1703.00548}.

\bibitem[\protect\citeauthoryear{Mnih \bgroup et al\mbox.\egroup
  }{2015}]{Mnih2015}
Mnih, V.; Kavukcuoglu, K.; Silver, D.; Rusu, A.~A.; Veness, J.; Bellemare,
  M.~G.; Graves, A.; Riedmiller, M.; Fidjeland, A.~K.; Ostrovski, G.; et~al.
\newblock 2015.
\newblock Human-level control through deep reinforcement learning.
\newblock {\em Nature} 518(7540):529--533.

\bibitem[\protect\citeauthoryear{Mobahi}{2016}]{Mobahi2016}
Mobahi, H.
\newblock 2016.
\newblock Training recurrent neural networks by diffusion.
\newblock {\em arXiv preprint arXiv:1601.04114}.

\bibitem[\protect\citeauthoryear{Naddef and Rinaldi}{2001}]{Naddef2001}
Naddef, D., and Rinaldi, G.
\newblock 2001.
\newblock The vehicle routing problem.
\newblock Philadelphia, PA, USA: Society for Industrial and Applied
  Mathematics.
\newblock chapter Branch-and-cut Algorithms for the Capacitated VRP,  53--84.

\bibitem[\protect\citeauthoryear{Penny and Sengupta}{2016}]{Penny2016}
Penny, W., and Sengupta, B.
\newblock 2016.
\newblock Annealed importance sampling for neural mass models.
\newblock {\em PLoS computational biology} 12(3):e1004797.

\bibitem[\protect\citeauthoryear{Sundar \bgroup et al\mbox.\egroup
  }{2012}]{Sundar2012}
Sundar, H.; Biros, G.; Burstedde, C.; Rudi, J.; Ghattas, O.; and Stadler, G.
\newblock 2012.
\newblock Parallel geometric-algebraic multigrid on unstructured forests of
  octrees.
\newblock In {\em Proceedings of the International Conference on High
  Performance Computing, Networking, Storage and Analysis}, ~43.
\newblock IEEE Computer Society Press.

\bibitem[\protect\citeauthoryear{Sutskever, Vinyals, and
  Le}{2014}]{Sutskever2014}
Sutskever, I.; Vinyals, O.; and Le, Q.~V.
\newblock 2014.
\newblock Sequence to sequence learning with neural networks.
\newblock In {\em Advances in neural information processing systems},
  3104--3112.

\bibitem[\protect\citeauthoryear{Vese}{1999}]{Vese1999}
Vese, L.
\newblock 1999.
\newblock A method to convexify functions via curve evolution.
\newblock {\em Communications in partial differential equations}
  24(9-10):1573--1591.

\bibitem[\protect\citeauthoryear{Vinyals and Le}{2015}]{Vinyals2015}
Vinyals, O., and Le, Q.
\newblock 2015.
\newblock A neural conversational model.
\newblock {\em arXiv preprint arXiv:1506.05869}.

\bibitem[\protect\citeauthoryear{Wang \bgroup et al\mbox.\egroup
  }{2015}]{Wang2015}
Wang, Z.; Schaul, T.; Hessel, M.; Van~Hasselt, H.; Lanctot, M.; and De~Freitas,
  N.
\newblock 2015.
\newblock Dueling network architectures for deep reinforcement learning.
\newblock {\em arXiv preprint arXiv:1511.06581}.

\bibitem[\protect\citeauthoryear{Ziebart \bgroup et al\mbox.\egroup
  }{2008}]{Ziebart2008}
Ziebart, B.~D.; Maas, A.~L.; Bagnell, J.~A.; and Dey, A.~K.
\newblock 2008.
\newblock Maximum entropy inverse reinforcement learning.
\newblock In {\em AAAI}, volume~8,  1433--1438.
\newblock Chicago, IL, USA.

\end{thebibliography}
\bibliographystyle{aaai}

\end{document}